%% file: main.tex
\documentclass[letterpaper, 10 pt, conference]{ieeeconf}  

\IEEEoverridecommandlockouts                              

\input{0-macro.tex}

\title{\LARGE \bf
Discrete Policy: Learning Disentangled Action Space \\ for Multi-Task Robotic Manipulation
}

\author{Kun Wu$^{1,*}$, Yichen Zhu$^{2,*}$, Jinming Li$^{3}$, Junjie Wen$^{4}$, \\Ning Liu$^{5}$, Zhiyuan Xu$^{5}$, and Jian Tang$^{5,\dagger}$
\thanks{$^{1}$Syracuse University, NY, USA {\tt\small kwu102@syr.edu}}
\thanks{$^{2}$Midea Group, AI Research Center, China {\tt\small zhuyc25@midea.com}}
\thanks{$^{3}$Shanghai University, China
        {\tt\small ljm2022@shu.edu.cn}}
\thanks{$^{4}$East China Normal University, China
        {\tt\small jjwen@cs.ecnu.edu.cn}}
\thanks{$^{5}$Beijing Innovation Center of Humanoid Robotics, Beijing, China
        {\tt\small \{Neil.Liu, Eric.Xu, jian.tang\}@x-humanoid.com }}
\thanks{$^*$ Co-first author. $\dagger$ Corresponding author: Jian Tang.}
}

\begin{document}

\makeatletter
\let\@oldmaketitle\@maketitle%
\renewcommand{\@maketitle}{\@oldmaketitle%
    \centering
    \includegraphics[width=0.8\linewidth]{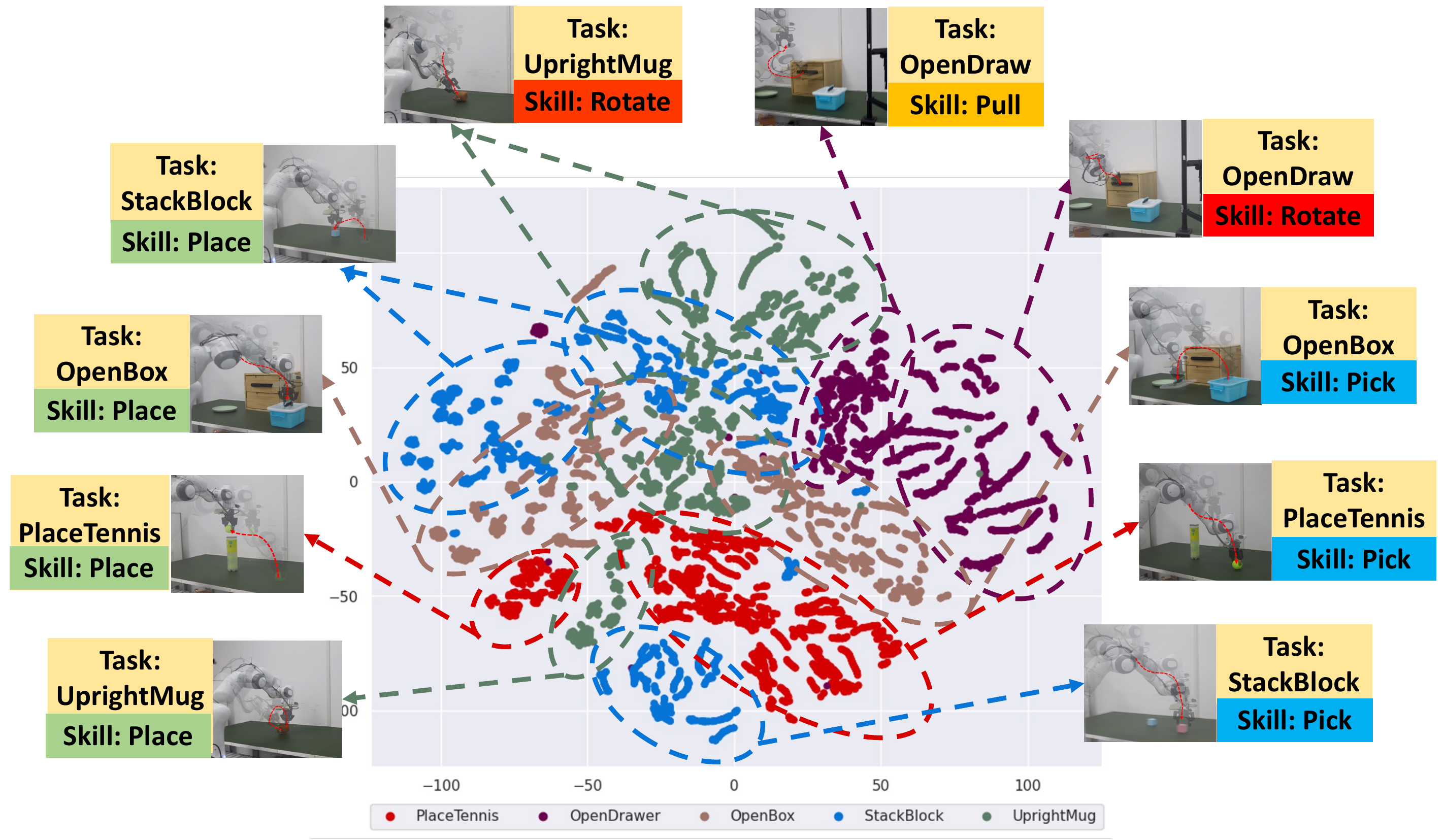}
    \captionof{figure}{\textbf{Visualization of Discrete Policy.} The t-SNE visualization of feature embeddings from Discrete Policy reveals that skills across different tasks cluster closely together. This pattern suggests that discrete latent spaces are capable of disentangling the complex, multimodal action distributions encountered in multi-task policy learning.}
    \label{fig:motivation}
}
\makeatother

\maketitle

\thispagestyle{empty}
\pagestyle{empty}


\begin{abstract}

\input{1-abstract.tex}

\end{abstract}


\input{2-intro}

\input{3-related}

\input{4-preliminary}

\input{5-method}

\input{6-exp}

\input{7-conclusion}




\bibliographystyle{IEEEtran}
\bibliography{reference}

\end{document}

%% file: 0-macro.tex
\usepackage{makecell} 
\usepackage{stfloats}                                   
\usepackage[misc]{ifsym}     
\usepackage{hyperref}   

\usepackage{booktabs}                                   
\usepackage{multirow}                                   
\usepackage{tablefootnote}                              
\usepackage[symbol]{footmisc}                           
\usepackage{amsmath,amssymb}                            
\usepackage{xcolor}                                     
\let\labelindent\relax
\usepackage{enumitem}                                   
\usepackage{subcaption}                                 
\usepackage{caption}     

\usepackage{graphicx}
\usepackage{cleveref}
\usepackage{csquotes}

\newcommand{\field}[1]{\mathbb{#1}}

\newcommand{\mat}[1]{\boldsymbol{#1}}                   

\newcommand{\eat}[1]{}                                  

\usepackage{cite}
\usepackage{nicematrix}
\usepackage{bm}
\usepackage{pifont}

\newcommand{\ourmethod}{{Discrete Policy}}
\newcommand{\methodname}{{Discrete Policy}}
\newcommand{\R}{\field{R}}

\newcommand{\bobs}      {\mat{o}}
\newcommand{\blang}      {\mat{l}}
\newcommand{\bcode}      {\mat{e}}
\newcommand{\bcamera}   {\mat{c}}
\newcommand{\bmea}      {\mat{m}}

\newcommand{\maL}        {\mathcal{L}}



%% file: 1-abstract.tex
Learning visuomotor policy for multi-task robotic manipulation has been a long-standing challenge for the robotics community. The difficulty lies in the diversity of action space: typically, a goal can be accomplished in multiple ways, resulting in a multimodal action distribution for a single task. The complexity of action distribution escalates as the number of tasks increases. In this work, we propose \textbf{Discrete Policy}, a robot learning method for training universal agents capable of multi-task manipulation skills. 
{\methodname} employs vector quantization to map action sequences into a discrete latent space, facilitating the learning of task-specific codes. These codes are then reconstructed into the action space conditioned on observations and language instruction. We evaluate our method on both simulation and multiple real-world embodiments, including both single-arm and bimanual robot settings. We demonstrate that our proposed {\methodname} outperforms a well-established Diffusion Policy baseline and many state-of-the-art approaches, including ACT, Octo, and OpenVLA. For example, in a real-world multi-task training setting with five tasks, {\methodname} achieves an average success rate that is 26\% higher than Diffusion Policy and 15\% higher than OpenVLA. As the number of tasks increases to 12, the performance gap between Discrete Policy and Diffusion Policy widens to 32.5\%, further showcasing the advantages of our approach. Our work empirically demonstrates that learning multi-task policies within the latent space is a vital step toward achieving general-purpose agents.
Our project is at \href{https://discretepolicy.github.io}{https://discretepolicy.github.io}.

%% file: 2-intro.tex
\section{Introduction}
In the realm of robotics, the ability to train robots for multi-tasking operations presents significant challenges that stem from the inherent complexity of handling diverse tasks simultaneously~\cite{bousmalis2023robocat}. Traditional robotic systems often focus on specialized tasks, but the dynamic environments in which modern robots operate demand versatile functionalities that can adapt to various situations.

However, the unique challenges of predicting robotic actions — such as multimodal action distributions, sequential correlations, the need for high precision, and noise in expert demonstration data — make learning a single task through imitation learning a formidable problem. When extending this approach to multi-task imitation learning, these challenges intensify. The tasks may involve a mixture of short-, medium-, and long-horizon objectives of varying lengths, further complicating the action space. This complexity results in an entangled, more diverse multimodal action distribution as the characteristics of single tasks merge. This fusion exacerbates the difficulty of accurately learning and executing multiple tasks simultaneously. 
Naive imitation learning, such as Behavior Cloning~\cite{florence2022implicit} and Diffusion Policy~\cite{chi2023diffusion_policy}, directly maps the observation and instruction to the action space. This mapping operates on the task-entangled dimension, which makes it hard for the policy network to distinguish different tasks from them, not to mention learning precise behavior.

In this work, we present Discrete Policy, a framework for multi-task visuomotor policy learning. Discrete Policy consists of two components: a vector-quantized variation autoencoder (VQ-VAE) and a conditional diffusion model. During training, the VQ-VAE is used to project the action sequence into discrete latent space and extract the 
discrete action modal abstraction $z$ for the current task.
This $z$ represents the latent feature embedding of the current task. 
Then, a conditional diffusion model is leveraged to conduct the denoising process in the discrete latent space. Finally, the task-specific latent embedding is mapped to the decoder of VQ-VAE to reconstruct the action space, conditioned on the observation and language instruction.

To illustrate our method, we present a visualization that employs t-SNE~\cite{vandermaaten08tsne} to demonstrate the feature embedding of Discrete Policy on training with five real-world tasks, which is shown in Figure~\ref{fig:motivation}. 
Intriguingly, we find that tasks with similar characteristics are located in proximity to each other. 
For example, the feature embeddings on the left primarily represent the skills associated with \enquote{placing}, while those at the bottom correspond to \enquote{picking} skills. 
These skills, which appear across various tasks, have closely situated feature embeddings, yet they still maintain discernible boundaries between them. Our observations suggest that a discrete latent space may be more effective for disentangling tasks and skills in policy learning. We conducted experiments across simulation and real-world data, including 23 tasks from RLBench, 12 tasks from single-arm robots, and 5 tasks from bimanual robots. These task types range from simple object pick-and-place operations to more complex, contact-rich manipulations like placing objects into a drawer. Our results demonstrate that our method, {\ourmethod}, significantly outperforms Diffusion Policy~\cite{chi2023diffusion_policy}, a strong baseline, by a considerable margin. Notably, the performance gap between {\ourmethod} and Diffusion Policy widens as the number of tasks incorporated into the training increases. Our approach also shows superior performance over multiple state-of-the-art methods including MT-ACT~\cite{bharadhwaj2023roboagent}, Octo~\cite{team2024octo}, and OpenVLA~\cite{kim24openvla}.

%% file: 3-related.tex
\section{Related Work}

\textbf{Diffusion model for policy learning.} Diffusion models belong to generative models. It progressively transforms random noise into a data sample, which has achieved great success in generating high-fidelity image and image editing~\cite{DDPM,DDIM,stablediffusion,Imagen}. 
Recently, diffusion models have been applied in robotics, combining with reinforcement learning~\cite{yang2023policy,mazoure2023value,brehmer2024edgi,venkatraman2023reasoning,lee2024refining,zhou2024adaptive,chen2022offline}, imitation
learning~\cite{chi2023diffusion_policy,prasad2024consistency,khazatsky2024droid,vosylius2024render, zhu2024retrieval,reuss2024multimodal,ze20243d,team2024octo,xian2023unifying,ze2023gnfactor,ha2023scaling,reuss2023goal,pearce2023imitating, moedp, mail, zhu2024scaling, wen2024tinyvla,wen2024diffusion,wen2025dexvla}, reward learning~\cite{psenka2023learning,huang2023diffusion}, grasping~\cite{urain2023se,simeonov2023shelving, wu2024learning}, and motion planning~\cite{zhong2023language,urain2023se,liu2022structdiffusion,saha2023edmp,chang2023denoising,zhu2024language,wen2024object}. In this work, we leverage the diffusion model to perform noising-denoising on the discrete latent space, instead of high-dimensional feature space as conventional approaches do.

\textbf{Multi-tasking in robotics.} Multi-task robotic manipulation has led to significant progress in the execution of complex tasks and the ability to generalize to new scenarios~\cite{brohan2022rt1, qtopt, shridhar2022cliport, ze2023gnfactor, ze2023visual, ze20243d, zhaoaloha, aldaco2024aloha, li2024llara, niu2024llarva}. Notable methods often involve the use of extensive interaction data~\cite{bcz, qtopt, shridhar2022cliport,wu2024robomind} including OpenX~\cite{o2024open} and RoboMIND~\cite{wu2024robomind} to train multi-task models. For example, RT-1~\cite{brohan2022rt1} underscores the benefits of task-agnostic training, demonstrating superior performance in real-world robotic tasks across a variety of datasets. RT-2~\cite{brohan2023rt-2} trains with mixed robot data and image-text pairs. PerAct~\cite{kumar2022pre} encodes language goals and shows its effectiveness in real robot experiments. Octo~\cite{team2024octo} uses cross-embodiment data for pertaining. This paper proposes a new approach to learning multi-task policy in the discrete latent space. 

\setcounter{figure}{1}
\begin{figure}[t]
    \centering
    \includegraphics[width=0.6\textwidth,trim=80 80 0 0,clip]{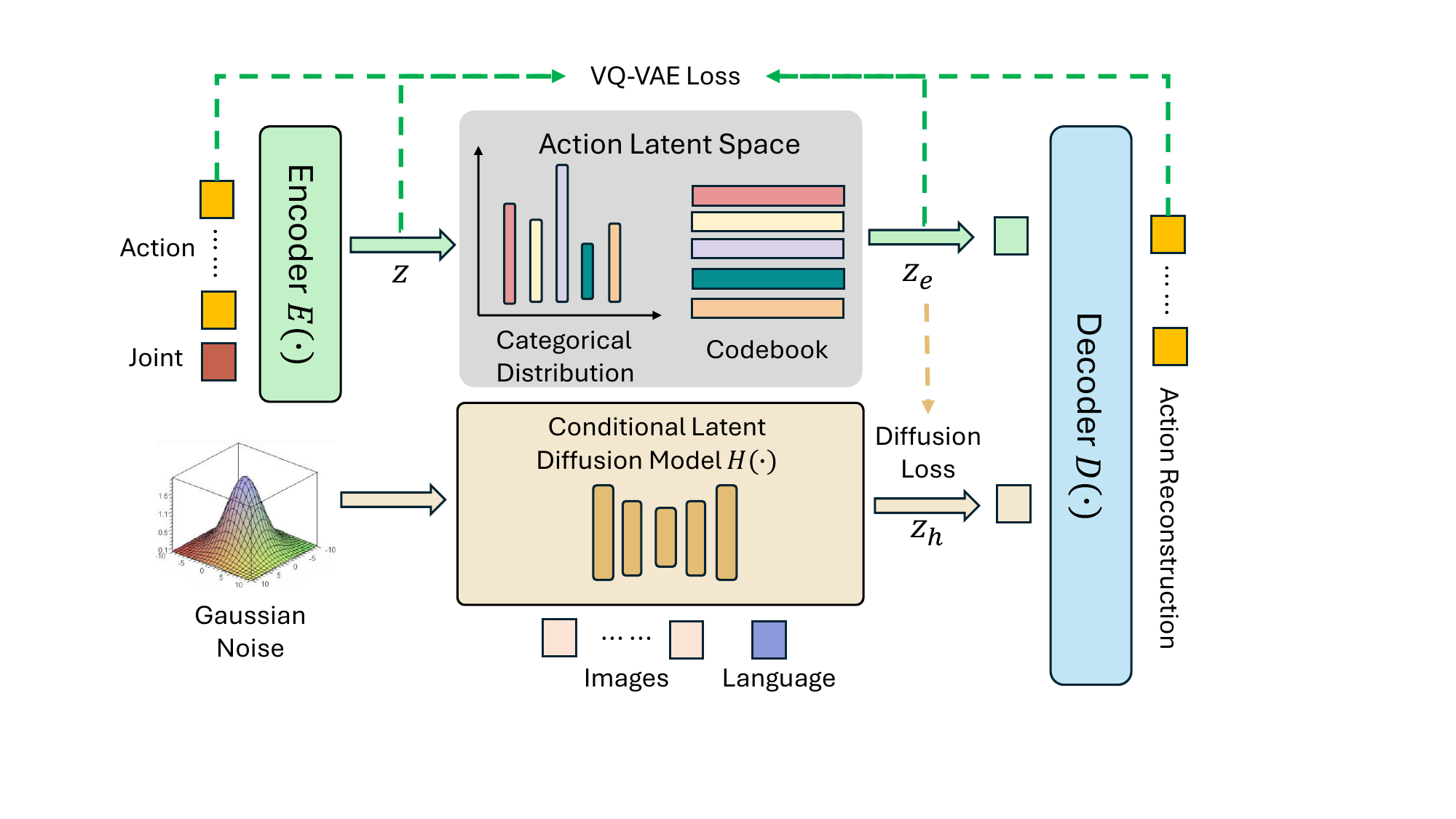}
    \caption{Overview of {\ourmethod}. In the first training stage, as indicated by the green arrow, we train a VQ-VAE that maps actions into discrete latent space with an encoder and then reconstructs the actions based on the latent embeddings using a decoder. In the second training stage, as indicated by the brown arrow, we train a latent diffusion model that predicts task-specific latent embeddings to guide the decoder in predicting accurate actions.}
    \label{fig:overview}
\end{figure}

%% file: 4-preliminary.tex
\section{Preliminaries}


Denoising Diffusion Probabilistic Models (DDPMs)~\cite{ho2020denoising}, as a class of expressive generative models, leverage Stochastic Langevin Dynamics~\cite{welling2011bayesian} to model the output generation process.
During the inference phase, starting from a Gaussian noise $x^K$, DDPMs perform $K$ iterations of the denoising process to generate a sequence of intermediate results $x^{k-1}, \cdots, x^{0}$ and take $x^{0}$ as the final denoised output.

To train the DDPMs, a denoising iteration $k$ is randomly selected for each data sample in the training dataset, and the corresponding noise $\epsilon^k$ is sampled from the noise scheduler. 
Then the noise prediction network $H(\cdot)$ is trained with the following objective:
\begin{align}
\label{Equation: pre_ddpm_train}
    \maL_{ddpm} = {|| \epsilon^k - H(x^{0}+\epsilon^k, k)  ||}_{2}.
\end{align}
In {\ourmethod}, we employ the Denoising Diffusion Implicit Models (DDIM)~\cite{song2020denoising} as the latent diffusion model, which improves upon DDPMs with faster inference speed.

%% file: 5-method.tex
\section{Methodology}

\input{5-1-overview}

\input{5-2-vqvae}

\input{5-3-ldp}

\input{5-4-implement}

%% file: 5-1-overview.tex
\subsection{Overview}

In this work, we introduce {\ourmethod}, designed to simultaneously learn multiple robotic manipulation tasks. 
As depicted in the Figure~\ref{fig:overview}, {\ourmethod} comprises two main components:
1) \textbf{Training stage 1}: We train a Vector Quantized Variational Autoencoder (VQ-VAE)~\cite{van2017neural} that encodes complex actions into a discrete latent space and subsequently reconstructs these actions using a decoder.
2) \textbf{Training stage 2}: We utilize a latent diffusion model to generate task-specific latent embeddings, which guide the decoder in executing the appropriate action modality.

\textbf{Formal definition.} At each inference timestamp $t$, {\ourmethod} take a language instruction $\blang$ and current observations $\bobs = \langle \bcamera, \bmea \rangle$ as input. 
The $\bcamera \in \R^{2 \times 3 \times H \times W}$ consists of two RGB images from the external fixed cameras on the left and right,
and $\bmea \in \R^{d}$ represents the low-dimensional proprioceptive states.
The latent diffusion model, conditioned on $\blang$ and $\bobs$, performs a denoising process to generate the latent action embedding $z$.
Subsequently, the VQ-VAE decoder uses the embedding $z$, along with $\blang$ and $\bobs$, to predict the final action $\hat{a}$.

%% file: 5-2-vqvae.tex
\subsection{Vector Quantized Autoencoder for Multimodal Action}

We detail the first training stage for VQ-VAE.
Unlike the vanilla Variational Autoencoder (VAE)~\cite{kingma2013auto}, which uses a continuous latent space and a Gaussian distribution prior, the VQ-VAE's discrete latent space simplifies the differentiation of skills and action modalities in robotic manipulation tasks. This facilitates more precise action predictions by the decoder.

To unify and compress different action modalities across tasks into a single latent space, the encoder $E(\cdot)$ processes the actions $a_{t:t+h}$ over a fixed horizon $h$ and the corresponding low-dimensional proprioceptive states $\bmea_{t:t+h}$, producing the latent action embedding $z = E(a_{t:t+h}, \bmea_{t:t+h})$.
We include the proprioceptive state in the input because it directly correlates with the robotic arm's actions, where  $a$ represents the 6D pose of the end effector, and $\bmea$ denotes the robotic joint positions in our implementation.
To ensure the encoder $E(\cdot)$ focuses on learning from the actions and not the auxiliary inputs, we do not include images $\bcamera$ or language instructions $\blang$ in the encoder's input.

To discretize the latent action space in the autoencoder, we maintain a codebook of discrete latent embedding $\bcode \in \R^{c \times f}$, where $c$ represents the number of the latent embedding categories and $f$ is the embedding dimension.
Given the latent action embedding $z$ from the encoder, the discrete latent embedding $z_{e}$ is selected through a bottleneck layer $S(\cdot)$ following the nearest neighbor lookup rule as follows:
\begin{align}
\label{Equation: vq_lookup}
    z_{e} = S(z) = e_{j}, {\rm where} \ j = argmin_{i} {|| z - e_{i} ||}_{2},
\end{align}
where $e_{i}$ is the discrete latent embedding for the $i$-th category in the codebook.
Then, the decoder $D(\cdot)$ takes the discrete latent embedding $z_{e}$ combined with the language instruction $\blang$ and current observations $\bobs$ to predict the action $\hat{a}$.
Similar to the loss function in VQ-VAE~\cite{van2017neural}, the final objective consists of three terms:
\begin{align}
\label{Equation: vq_loss}
    \maL_{vq} = {||\hat{a} - a||}_{1} + \beta {|| sg(z) - z_e ||}_{2}^{2} + \beta {|| z - sg(z_e) ||}_{2}^{2},
\end{align}
where $sg(\cdot)$ is the operator for stopping the gradient, the first term is to reconstruct the input, the second term is the quantization loss to train the codebook, and the third term is the commitment loss to encourage the encoder to commit to a code.
$\beta$ is the coefficient to balance the losses, which is set to 1.0 in our implementation

%% file: 5-3-ldp.tex
\subsection{Latent Diffusion Model}

After training the VQ-VAE, a naive sampling method is to use a uniform categorical distribution and select one latent embedding randomly from the codebook $e$.
However, for the multi-task robotic manipulation problem, this method risks selecting a prior latent embedding $z_e$ that is inappropriate for the current task and leads to task failure.
Therefore, we freeze the trained VQ-VAE and employ a conditional latent diffusion model $H(\cdot)$ to generate the discrete latent embedding $z_e$ suitable for the current task. 
Given the noisy latent embedding $z^k$ at the denoise iteration $k$ and the language instruction $\blang$ and current observations $\bobs$, the latent diffusion model $H(\cdot)$ outputs the corresponding noise for the task-specific latent embedding $z_h$.
Instead of directly predicting the latent embedding $z_h$, the latent diffusion model $H(\cdot)$ predicts the noise at each iteration and tries to recover the latent embedding $z_h$ from the Gaussian noise through multi-round denoising processes.
Following~\cite{khazatsky2024droid}, we use DDIM~\cite{song2021denoising} as the latent diffusion model to improve the inference efficiency and use the same noise scheduler and hyperparameters.
The training objective is as follows:
\begin{align}
\label{Equation: vq_loss}
    \maL_{ldm} = {|| \epsilon^k - H(z_{e}^{k}, \blang, \bobs, k)  ||}_{2},
\end{align}
where $k$ represents the $k$-th denoise iteration.
During the inference, starting from Gaussian noise, the latent diffusion model $H(\cdot)$ performs $K$ times denoise processes and recovers the latent embedding $z_h$.
Then the latent embedding $z_h$ is passed to the bottleneck layer $S(\cdot)$ in VQ-VAE for finding the most similar discrete latent embedding $z_e = S(z_h)$, and then the decoder $D(\cdot)$ predicts the final action $\hat{a}$ based on the discrete latent embedding $z_e$.

%% file: 5-4-implement.tex
\subsection{Network Architecture}
In our work, we leverage both language instruction and images to predict the robot's action.  In particular, for language instructions, we use a pre-trained DistilBERT~\cite{sanh2019distilbert} to extract the features. For RGB images, we use EfficientNet B3~\cite{tan2019efficientnet} as the visual backbone and fuse the linguistic and visual features by FiLM layers~\cite{perez2018film}. We also include proprioceptive states, we use MLP layers to project the input to the features with a fixed dimension of 512. We attached a transformer model to VQ-VAE to unify all input tokens into a single embedding. To be specific, the VQ-VAE is equipped with a transformer encoder with 4 layers, and a transformer decoder with 7 layers. The hidden dimension is set to 512. For the diffusion model which operates on the discrete latent space, we use Unet~\cite{ronneberger2015u} to keep the input (i.e., noisy latent embedding) and output (i.e., denoised latent embedding) dimensions identical.

%% file: 6-exp.tex
\section{Experiments}

After presenting {\ourmethod}, we ask the following key questions about the effectiveness of our algorithm:
1) Can {\ourmethod} be effectively deployed to real-world scenarios? 
2) Can {\ourmethod} be scaled up to multiple complex tasks? 
3) Can {\ourmethod} effectively distinguish different behavioral modalities across multiple tasks? 
In order to answer these questions, we built a real-world robotic arm environment, 
designed a variety of manipulation tasks that contain rich skill requirements and long-horizon tasks, 
and finally conducted extensive experiments as well as visualization of {\ourmethod}.

\begin{figure}[tp]
\centering
\includegraphics[width=0.9\linewidth,trim=35 0 0 0,clip]{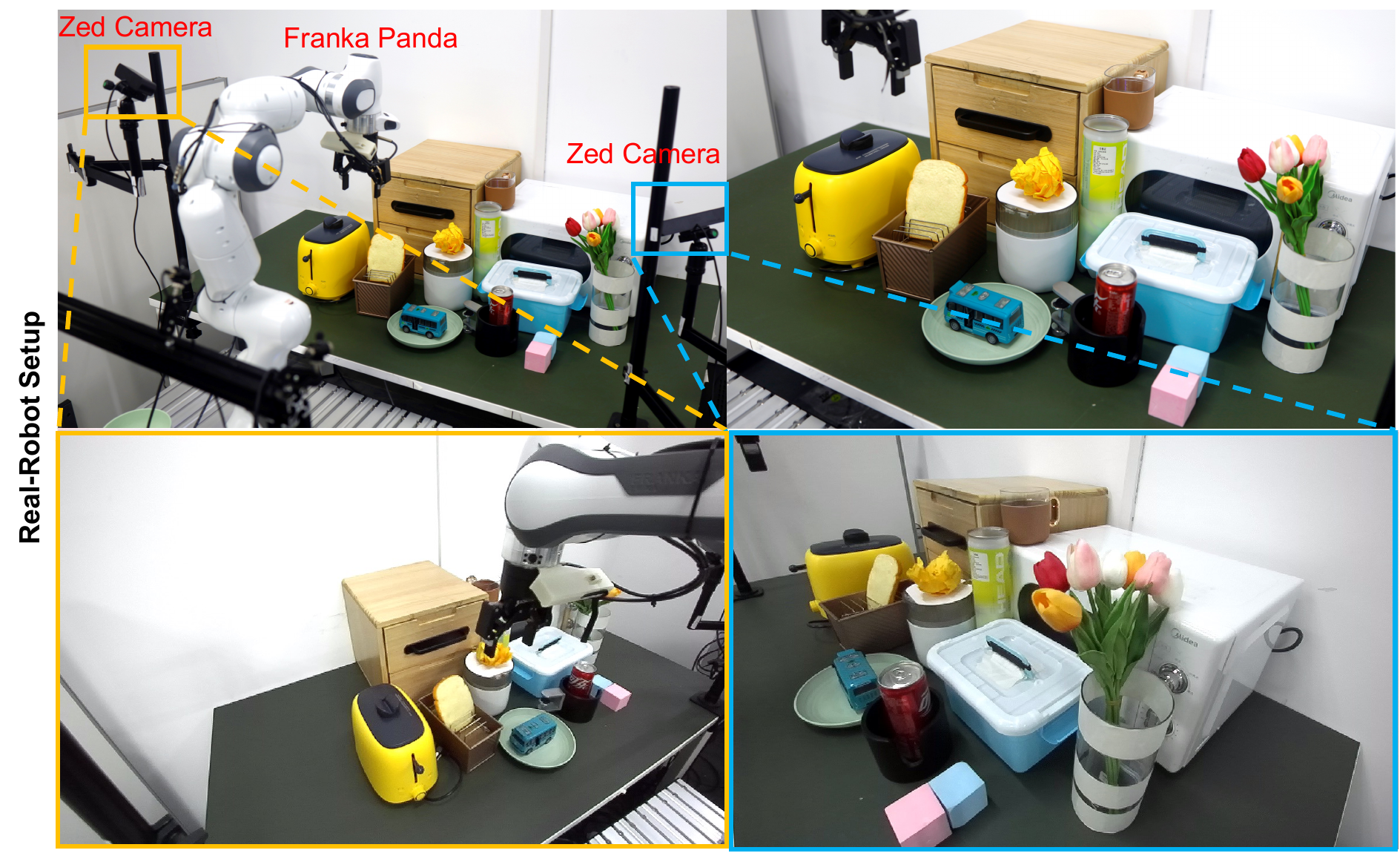}
\caption{Real-World Experiment Setup for single-arm Franka robot. We use two external fixed-view Zed cameras. The figure in the upper right corner shows all the objects used in our experiments.}
\label{fig:franka_real_world_setting}
\end{figure}

\begin{figure}[tp]
\centering
\includegraphics[width=0.9\linewidth,trim=35 0 0 0,clip]{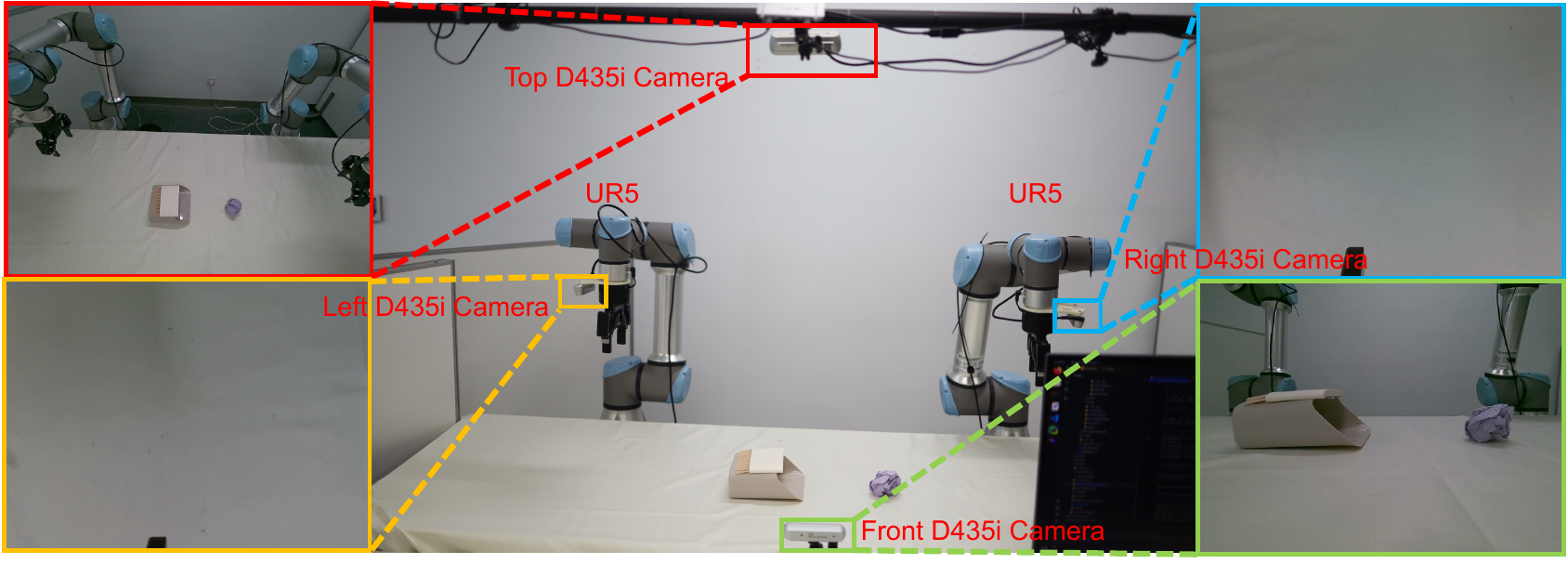}
\caption{Real-World Experiment Setup for bimanual UR5 robot. We use four fixed-view RealSense D435i cameras.}
\label{fig:ur5_real_world_setting}
\end{figure}

\begin{figure}[tp]
\centering
\includegraphics[width=0.98\linewidth,trim=0 130 0 130,clip]{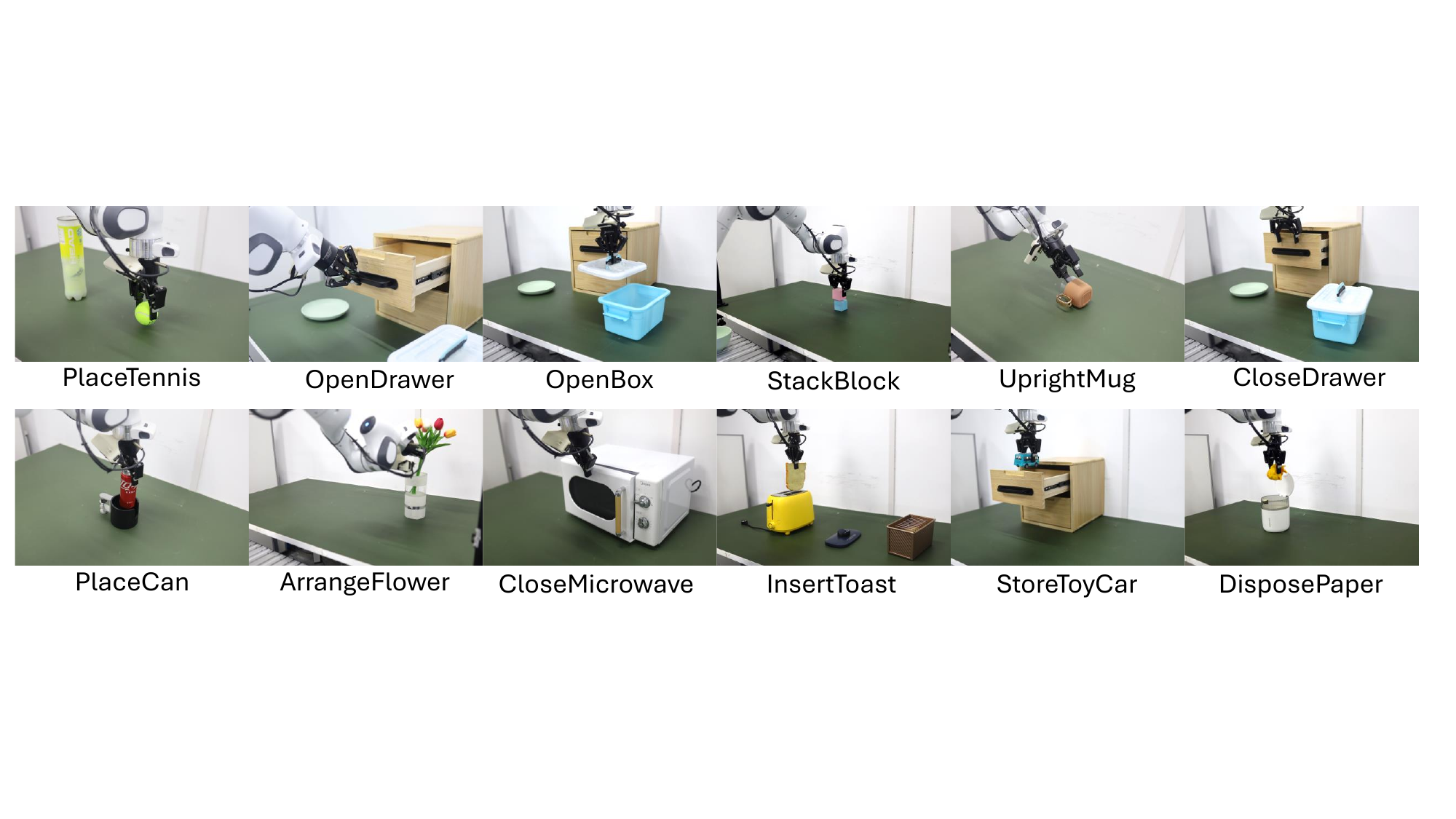}
\caption{Demonstrations of the 12 tasks in single-arm robot experiments.}
\label{fig:franka_all_task_demo}
\end{figure}

\begin{figure}[tp]
\centering
\includegraphics[width=0.98\linewidth,trim=0 10 0 10,clip]{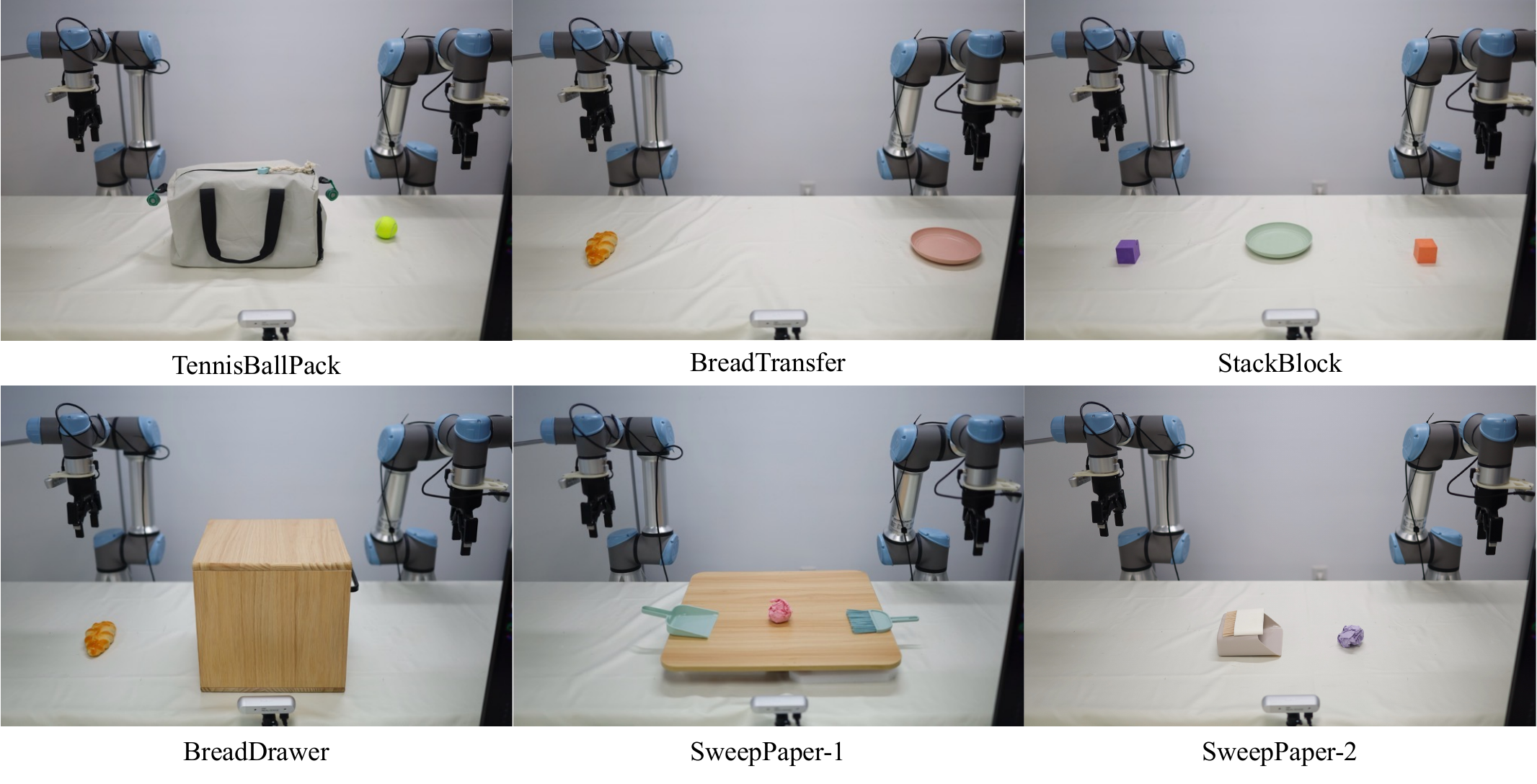}
\caption{Demonstrations of the 6 tasks in bimanual robot experiments.}
\label{fig:ur5_all_task_demo}
\end{figure}

\input{task_summary}

\input{6-1-setup}

\input{6-2-eval}

\input{6-3-ablation}
\input{6-4-visual}

%% file: task_summary.tex
\begin{table}[t]
  \centering
   \caption{The tasks summary of our real-world experiments.}
  \label{table:each_data_summary}
  \resizebox{\columnwidth}{!}{
      \begin{tabular}{cccccc}
      \toprule
      \multirow{2}{*}{\#} & \multirow{2}{*}{Task} & Horizon  & \# of  & Avg.  & \multirow{2}{*}{Task Instruction} \\
      &  & Type &  Demo. & Traj. Length  & \\
      \midrule
      \multicolumn{6}{c}{\textbf{Single-Arm Franka Robot}} \\
      \cmidrule{1-6}
      T1 & PlaceTennis & Short & 100 & 170.7 & Place the tennis ball in the tube. \\
      T2 & OpenDrawer & Short & 100 & 284.1 & Pull open the drawer. \\
      T3 & OpenBox & Short & 100 & 231.4 & Open the blue plastic box. \\
      T4 & StackBlock & Short & 100 & 229.1 & Stack the pink block on the blue block. \\
      T5 & UprightMug & Short & 100 & 195.4 & Turn the tipped mug upright. \\
      T6 & CloseDrawer & Short & 100 & 176.1 & Push the drawer closed. \\
      T7 & PlaceCan & Short & 100 & 237.9 & Put the drink can in the cup holder. \\
      T8 & ArrangeFlower & Short & 100 & 282.2 & Put the bouquet in the vase. \\
      T9 & CloseMicrowave & Short & 100 & 164.4 & Close the microwave door. \\
      T10 & InsertToast & Long & 100 & 475.9 & Place the toast in the toaster. \\
      T11 & StoreToyCar & Long & 100 & 457.2 & Put the toy car in the drawer. \\
      T12 & DisposePaper & Long & 100 & 295.2 & Put the trash in the bin. \\
      \midrule
      \multicolumn{6}{c}{\textbf{Bimanual UR5 Robot}} \\
      \midrule
      T1 & TennisBallPack & Long & 64 & 438.5 & Unzip the bag, place the tennis ball into the bag. \\
      T2 & BreadTransfer & Short & 100 & 250.0 & Place bread on the plate. \\
      T3 & StackBlock & Short  & 150 & 200.0 & Stack the orange cube on the purple cube. \\
      T4 & BreadDrawer & Long & 100 & 448.9 & Place the bread into the drawer. \\
      T5 & SweepPaper-1 & Long & 100 & 320.0 & Sweep trash into the green dustpan. \\
      T6 & SweepPaper-2 & Long & 100 & 300.0 & Sweep trash into the white dustpan. \\
      \bottomrule
      \end{tabular}
  }
\end{table}

%% file: 6-1-setup.tex
\subsection{Experiment Setup}

For real-world experiments, we collect the dataset through human demonstration. Given a target task, we randomly place the objects within a specified area and ask the human demonstrators to finish the task nearly perfectly. We record the RGB stream from multiple camera views and robot states, e.g., joint position, during the whole robot control process. Our model predicts the 6D pose, including position $(x,y,z)$ and rotation $(roll, pitch, yaw)$. For the single-arm Franka robot shown in Figure~\ref{fig:franka_real_world_setting}, we follow Droid~\cite{khazatsky2024droid}, with two external fixed-view Zed 2 stereo cameras, one on the left and one on the right. For bimanual UR5 robotic arms shown in Figure~\ref{fig:ur5_real_world_setting}, we used four RealSense D435i cameras in the environment, two of which are hand-eye cameras mounted at the wrists of each arm, and the other two are external cameras mounted above and in front of the operating table.

For the simulation benchmark, we evaluate on Bi-DexHands~\cite{bi-dexhands} and Metaworld~\cite{yu2020metaworld} Medium level and Hard level, following the settings in MWM~\cite{seo2023masked}. All experiments were trained with 20 demonstrations and evaluated with 3 seeds, and for each seed, the success rate was averaged over five different iterations.

\begin{figure}[tp]
\centering
\includegraphics[width=0.99\linewidth]{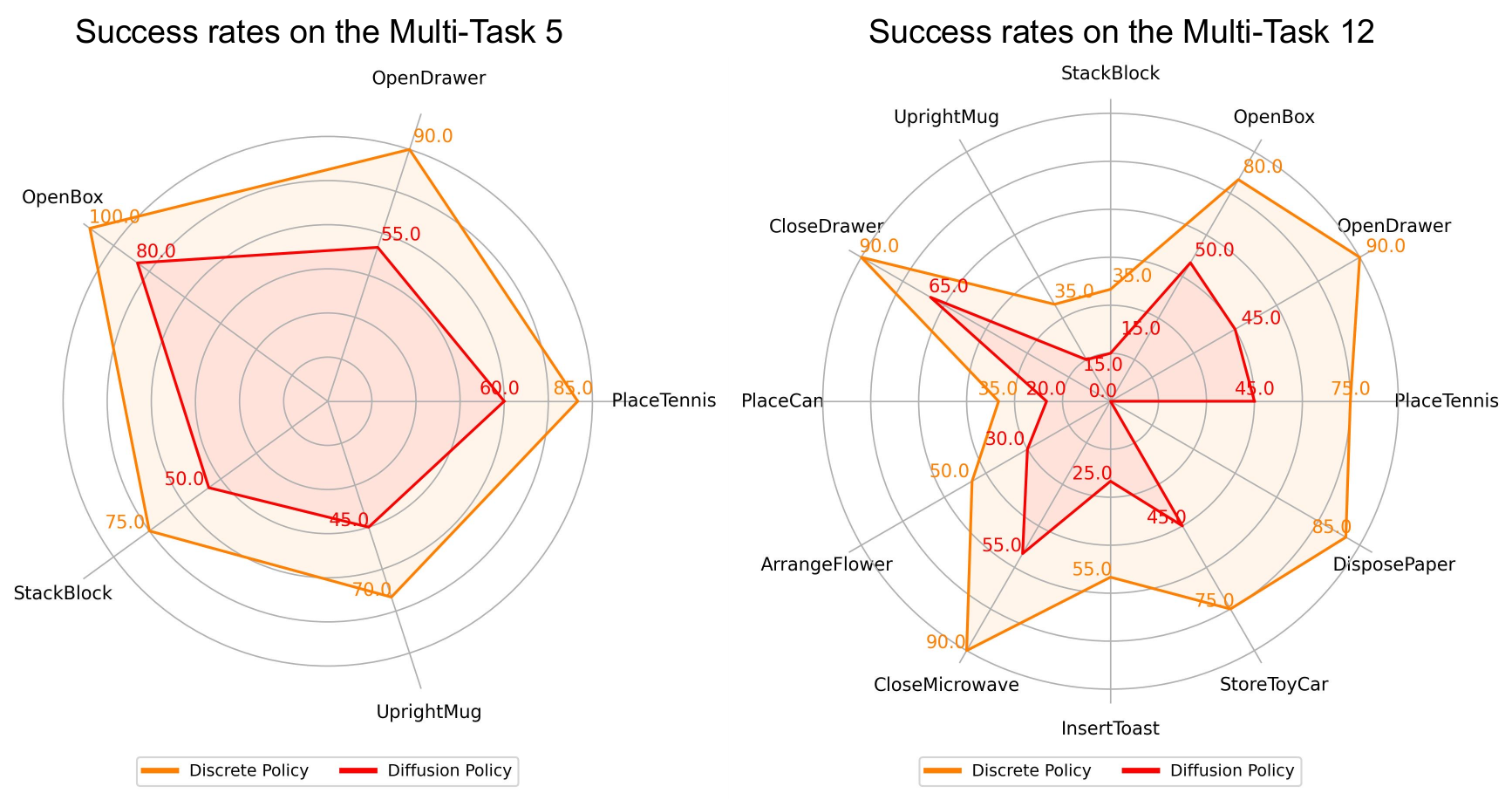}
\caption{The figures on the left and right show the success rates on the MT-5 and MT-12, respectively.}
\label{fig:main_result}
\end{figure}

\begin{table}[t]
  \centering
   \caption{Comparisons on \textbf{Multi-Task 5 (MT-5) in the single Franka robot}, with success rates reported and the best results highlighted in bold. The symbol * denotes that methods are pretrained by 970K OpenX~\cite{o2024open} robot data.}
  \label{table:mt5_franka_exp}
  \resizebox{\columnwidth}{!}{
      \begin{tabular}{c|ccccc|c}
      \toprule
      Method & Task1 & Task1 & Task3 & Task4 & Task5 & Average \\ 
      \midrule
      RT-1~\cite{brohan2022rt} & 25 & 30 & 30 & 10 & 15 & 22 \\
      BeT~\cite{shafiullah2022behavior} & 45 & 30 & 65 & 30 & 15 & 37 \\
      BESO~\cite{reuss2023goal} & 40 & 30 & 55 & 25 & 15 & 33 \\
      MDT~\cite{reuss2024multimodal} & 50 & 35 & 55 & 20 & 25 & 37 \\
      Octo*~\cite{octo_2023} & 40 & 55 & 50 & 30 & 40 & 43 \\
      OpenVLA*~\cite{kim24openvla} & \textbf{85} & 80 & 90 & 40 & 50 & 69 \\
      MT-ACT~\cite{bharadhwaj2023roboagent} & 80 & 80 & \textbf{100} & 55 & 45 & 59 \\ 
      Diffusion Policy~\cite{diffusion-policy} & 60 & 55 & 80 & 50 & 45 & 58\\
      \midrule
      Discrete Policy & \textbf{85} & \textbf{90} & \textbf{100} & \textbf{75} & \textbf{70} & \textbf{84} \\
      \bottomrule
      \end{tabular}
  }
\end{table}

\textbf{Task Description for Real-World Experiments.}
For the single-arm Franka experiments, we designed two multi-task evaluation protocols named Multi-Task 5 (MT-5) and Multi-Task 12 (MT-12).
MT-12 extends MT-5's task range to 12 tasks, including 3 long-horizon tasks, more varied scenarios, and more complex skill requirements such as flip, press, and pull, which are shown in Figure~\ref{fig:franka_all_task_demo}.
For the bimanual UR5 robotic experiments, as shown in Figure~\ref{fig:ur5_all_task_demo}, we designed 6 challenging tasks that require collaboration between two robotic arms.
We provide a summary of all the tasks in Table~\ref{table:each_data_summary}. For each task, we evaluate 20 times with different initial object states and report the success rates.

%% file: 6-2-eval.tex
\subsection{Experimental Results}

\textbf{Results on Single-arm Franka Robot.} Figure~\ref{fig:main_result} shows the comparisons of the success rates of all tasks on MT-5 and MT-12, respectively. We observe that {\ourmethod} achieves average success rates of 84\% on MT-5 and 66.3\% on MT-12, which significantly outperforms Diffusion Policy by 25\% and 32.5\%. Another observation is that the gap between average success rates becomes larger as more complex tasks evolve during training. We believe this is due to the increased number of tasks and the increased difficulty of the model distinguishing between individual tasks and learning the optimal policy among them. Notably, both methods suffer from a performance drop as the number of tasks increases. This is due to fixed total training iterations. 

In Table~\ref{table:mt5_franka_exp}, we compare our method with other state-of-the-art approaches, ranging from pure Transformer architectures trained using behavior cloning to diffusion-based policies. Notably, both OpenVLA and Octo are pre-trained on OpenX datasets, making a direct comparison with our method—trained solely on limited robot data—unfair. Despite this, our method consistently achieves the best performance and is on par with these methods for all tasks. In terms of average success rate, our approach even surpasses OpenVLA by 15\% over five tasks.

\input{bimanual_exp}

\textbf{Results on Bimanual UR5 robot.} We further conducted experiments using a bimanual UR5 robot. With an increased number of degrees of freedom, the complexity of the tasks rises significantly. In this setting, we trained all methods on six tasks, ranging from extremely long-horizon tasks like placing a tennis ball into a closed bag to more straightforward tasks such as transferring a piece of bread. The experimental results are illustrated in Table~\ref{tab:dualarm-mt-5_results}. We compare our approach to BeT~\cite{shafiullah2022behavior}, MT-ACT~\cite{bharadhwaj2023roboagent}, and Diffusion Policy~\cite{diffusion-policy}. Our approach achieves an average success rate exceeding 65\%, while Diffusion Policy achieves only 37.5\%. Our method also outperforms both MT-ACT and BeT. These empirical results demonstrate the effectiveness of our method in multi-task scenarios, even in the challenging bimanual manipulation setting.

\textbf{Results on Simulation.} Finally, we evaluated our approach on simulation benchmarks, selecting several complex tasks, including 6 tasks from Bi-DexHands~\cite{bi-dexhands}, 11 tasks from Metaworld Medium~\cite{yu2020metaworld}, and 6 tasks from Metaworld Hard~\cite{yu2020metaworld, seo2023masked}. The experiments are shown in Table~\ref{tab:simulation}. Compared to state-of-the-art methods, our approach demonstrates superior performance across all settings, further validating its effectiveness.

%% file: bimanual_exp.tex
\begin{table}[t]
    \centering
    \caption{Comparing Discrete Policy with baseline methods on \textbf{six bimanual UR5 robot} tasks.}
    \label{tab:dualarm-mt-5_results}
    \resizebox{0.5\textwidth}{!}{\begin{tabular}{c|cccccc|c}
    \toprule
    Method / Task & T1 & T2 & T3 &T4 &  T5 & T6 & Avg. \\ 
    \midrule
    BeT~\cite{shafiullah2022behavior} & 30 & 10 & 0 & 30 & 40 & 20  & 21.7 \\
    MT-ACT~\cite{bharadhwaj2023roboagent} & \textbf{70} & 40 & 10 & 70 & 80 & 60 & 55.0\\
    Diffusion Policy~\cite{diffusion-policy} & 30 & 35 & 0 & 45 & 65 & 50 & 37.5 \\
    \midrule
    Discrete Policy & \textbf{70} & \textbf{55} & \textbf{30} & \textbf{85} & \textbf{85} & \textbf{75} & \textbf{65.8} \\
    \bottomrule
    \end{tabular}}
\end{table}

\input{sim_exp}

%% file: sim_exp.tex
\begin{table}[t]
    \centering
    \caption{Experimental results of Bi-Dexhands~\cite{bi-dexhands} and Metaworld~\cite{yu2020metaworld}, a simulation benchmark. The numbers in parentheses indicate the number of tasks for the simulation benchmark.}
    \label{tab:simulation}
    \resizebox{0.45\textwidth}{!}{\begin{tabular}{c|ccc}
        \toprule
        \multirow{2}{*}{Method} & \multirow{2}{*}{\shortstack{Bi-Dex-\\Hands(6)}} & \multirow{2}{*}{\shortstack{Metaworld-\\Medium(11)}} & \multirow{2}{*}{\shortstack{Metaworld-\\Hard(6)}} \\
        &  &  &  \\
        \midrule
        VINN~\cite{parisi2022unsurprising} & 20.6 & 5.2 & 2.7 \\
        BeT~\cite{shafiullah2022behavior} &24.5 & 9.1 & 0.9 \\
        MT-ACT~\cite{bharadhwaj2023roboagent} &47.6 & 15.4 & 4.8 \\
        Diffusion Policy~\cite{diffusion-policy} & 35.7 & 16.2 & 3.1 \\
        \midrule
        Discrete Policy & \textbf{54.3} & \textbf{19.6} & \textbf{7.9} \\
        \bottomrule
    \end{tabular}}
\end{table}

%% file: 6-4-visual.tex
\subsection{Visulization on Action Space}

\begin{figure}[tp]
\centering
\includegraphics[width=\linewidth,trim=10 0 0 0,clip]{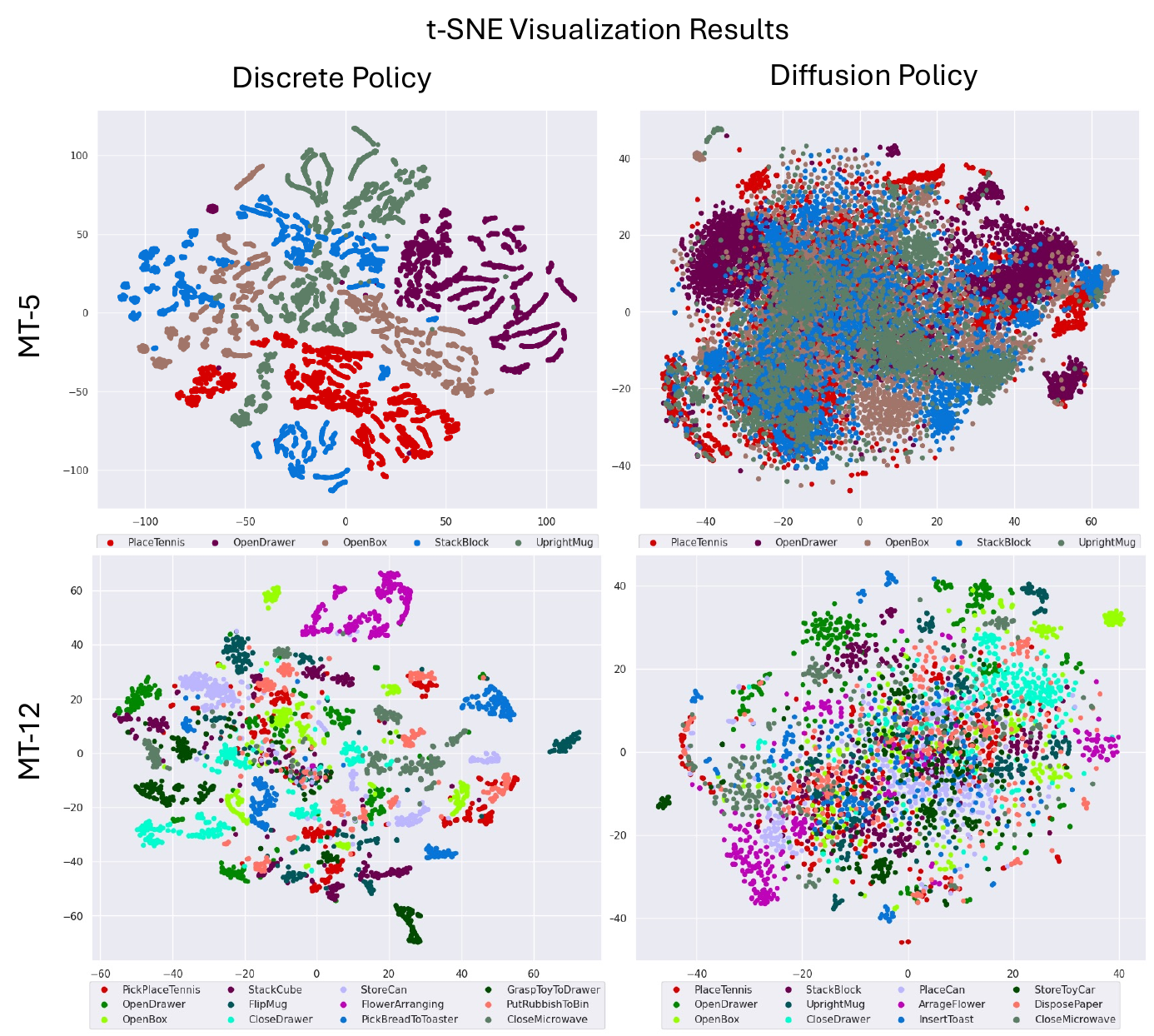}
\caption{The t-SNE visualization results of the latent features of {\ourmethod} and Diffusion Policy on MT-5 and MT-12.}
\label{fig:mt5-12_visual}
\end{figure}

To illustrate that {\ourmethod} can effectively disentangle action space between diverse tasks, we visualized the features in the latent space using the t-SNE~\cite{vandermaaten08tsne} in~\Cref{fig:motivation,fig:mt5-12_visual}. We compare our approach to the Diffusion Policy. For MT-5, we observe that {\ourmethod} clearly delineates different features while aggregating similar ones, indicating that {\ourmethod} effectively distinguishes between multimodal actions and skills across multiple tasks. In contrast, the features of Diffusion Policy are fragmented and overlap a lot.
In the MT-12 scenario, where the number of tasks is greater, delineating the distribution of actions becomes more challenging. Nonetheless, {\ourmethod} successfully distinguishes the action distributions in most tasks. Conversely, the complex action distributions confound the Diffusion Policy.

\subsection{Ablation Study}
We conducted extensive ablation studies to evaluate the effects of various hyperparameters.
The results of these experiments are summarized in Table~\ref{table:rb_ablation}.
For action chunk size, we observed that success rates increase as the chunk size grows, particularly when increasing from 16 to 32. We also experimented with different numbers of codebook categories to assess whether increasing them from 256 to 1024 would impact model performance. Our results show that increasing the number of codebook categories consistently improves the average success rates across all tasks. This improvement is likely due to the larger capacity of the codebook, which allows it to capture a broader range of behavioral patterns.

Latent embeddings play a crucial role in our method. We evaluate both the number and dimensionality of these embeddings. Empirically, we found that increasing the dimensionality of the latent embeddings enhances performance, leading to higher success rates. However, increasing the number of latent embeddings does not necessarily yield better results. Our observations indicate that setting the number of latent embeddings to eight provides optimal performance.

\begin{table}[t]
  \centering
   \caption{Ablation study on Multi-Task 5 (MT-5) in the single Franka robot environment in terms of the success rates with the best results in bold. }
  \label{table:rb_ablation}
  \resizebox{\columnwidth}{!}{
      \begin{tabular}{c|c|ccccc}
      \toprule
      Factor & Value & Task1 & Task2 & Task3 & Task4 & Task5 \\
      \midrule
      \multirow{2}{*}{Chunk Size} & 16 & 70 & \textbf{85} & \textbf{100} & 60 & 50 \\
        & 32  & \textbf{85} & \textbf{85} & \textbf{100} & \textbf{75} & \textbf{60} \\
      \midrule
      \multirow{3}{*}{Codebook Category}&  256 & 60 & 80 & 90 & 65 & 50 \\
       &  512  & 60 & \textbf{90} & 90 & 65 & 55 \\
       &1024  & \textbf{80} & \textbf{90} & \textbf{100} & \textbf{70} & \textbf{60} \\
      \midrule
      \multirow{3}{*}{Latent Embed. Dim.}& 32  & 50 & 75 & 85 & 50 & 55 \\
       & 64  & 60 & 75 & \textbf{95} & \textbf{70} & 65 \\
       &  128  & \textbf{80} & \textbf{90} & \textbf{95} & \textbf{70} & \textbf{70} \\
      \midrule
      \multirow{3}{*}{Latent Embed. Num.} &  1 & 60 & 85 & 95 & 65 & 55 \\
       &  8 & \textbf{80} & \textbf{90} & \textbf{100} & \textbf{75} & \textbf{75} \\
       & 16 & \textbf{80} & \textbf{90} & \textbf{100} & 70 & 70 \\
      \bottomrule
      \end{tabular}
  }
\end{table}

\begin{figure}[t]
\centering
\includegraphics[width=\linewidth,trim=0 0 0 0,clip]{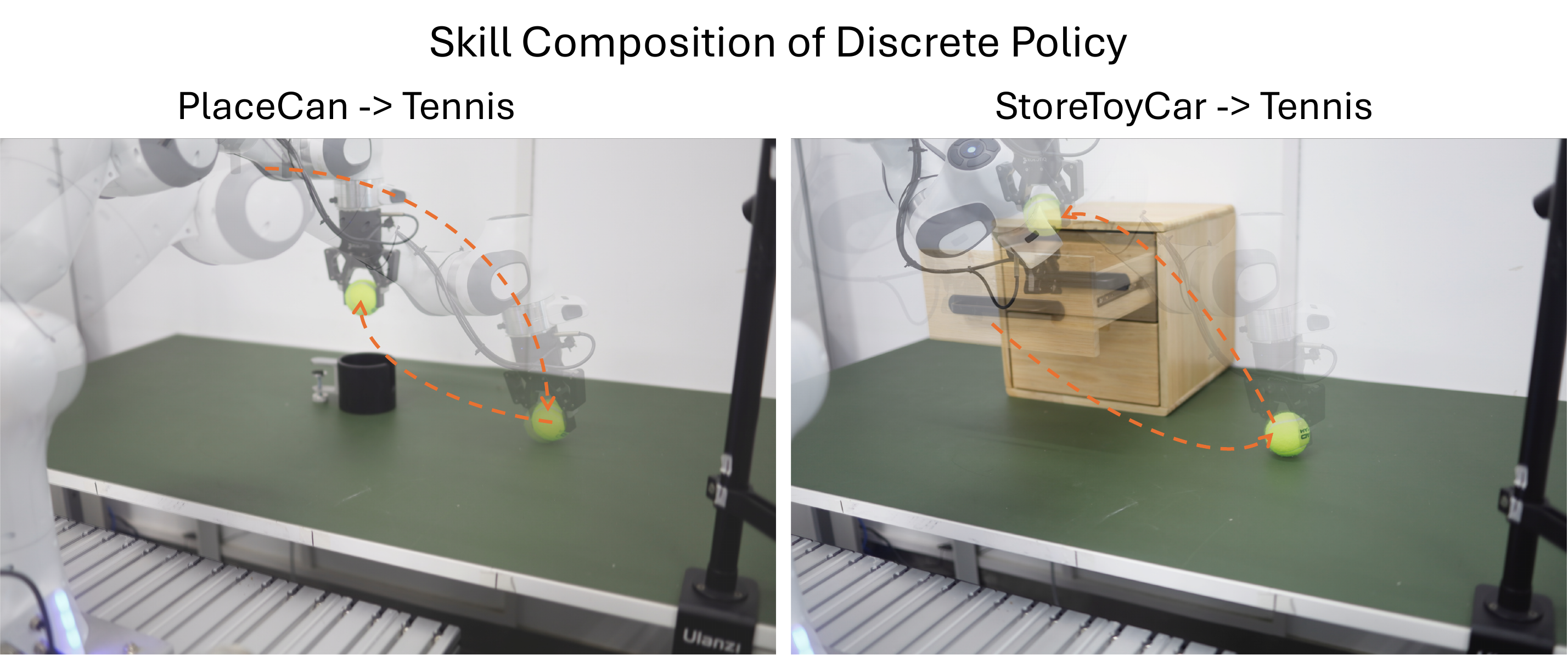}
\caption{Skill Composition of Discrete Policy.}
\label{fig:skill_compos}
\end{figure}

\subsection{Skill Composition}
Skill composition allows the model to map language queries to specific action spaces. By decomposing the action sequence into discrete latent embeddings, our Discrete Policy aims to align language inputs with this latent space~\cite{de2023cross, pmlr-v243-lahner24a}. 
To evaluate whether our method is capable of this, we combined two instructions from the training data and assessed whether the model could successfully complete the newly composed task.

As shown in Figure~\ref{fig:skill_compos}, we tasked the robot with two new objectives: 1) placing a tennis ball into a cup holder, and 2) placing a tennis ball into a drawer. Our Discrete Policy successfully interpreted these new instructions and completed the tasks, demonstrating the effectiveness of using discrete latent embeddings in action generation. This further supports the motivation behind our work, where a disentangled action space can be effectively composed into meaningful sequences.

%% file: 7-conclusion.tex
\section{Conclusion}
This paper explores innovative strategies for multi-task learning in robotic systems. Our method, called Discrete Policy, learns action patterns in the latent space, enabling better disentanglement of feature representations across different skills. Through extensive simulations and real-world experiments, our approach demonstrates superior performance in multi-task settings compared to various state-of-the-art methods. Overall, we believe that the Discrete Policy approach offers a compelling and practical new perspective on learning multi-task policies for embodied control.